\documentclass[letterpaper, 10 pt, conference]{ieeeconf}
\IEEEoverridecommandlockouts
\usepackage[pdftex]{graphicx}
\usepackage[tight]{units}
\usepackage{booktabs}
\usepackage{amsmath}
\usepackage{amsfonts}
\usepackage{algpseudocode}
\usepackage{algorithm}
\usepackage{color}
\usepackage[usenames,dvipsnames]{xcolor}
\newcommand\eat[1]{}

\begin{document}
\title{\LARGE \bf On-line and On-board Planning and Perception for Quadrupedal
Locomotion}
\author{{\centering 
Carlos Mastalli,~~Ioannis Havoutis,~~Alexander W. Winkler,~~Darwin G.
Caldwell,~~Claudio Semini} \\
Department of Advanced Robotics, Istituto Italiano di Tecnologia,
via Morego, 30, 16163 Genova, Italy
\thanks{\textit{email}: \{carlos.mastalli, ioannis.havoutis,
darwin.caldwell, claudio.semini\}@iit.it, alexander.winkler@mavt.ethz.ch}
}

\maketitle

\begin{abstract}
We present a legged motion planning approach for quadrupedal locomotion over 
challenging terrain. We decompose the problem into body action planning and 
footstep planning. We use a lattice representation together with a set of
defined body movement primitives for computing a body action plan. The lattice
representation allows us to plan versatile movements that ensure feasibility for
every possible plan. To this end, we propose a set of rules that define the
footstep search regions and footstep sequence given a body action. We use
Anytime Repairing A* (ARA*) search that guarantees bounded sub-optimal plans.
Our main contribution is a planning approach that generates on-line versatile
movements. Experimental trials demonstrate the performance of our planning
approach in a set of challenging terrain conditions. The terrain information and
plans are computed on-line and on-board.
\end{abstract}

\section{Introduction}
Legged motion planning over rough terrain involves making careful decisions
about the footstep sequence, body movements, and locomotion behaviors. Moreover,
it should consider whole-body dynamics, locomotion stability, kinematic and
dynamic capabilities of the robot, mechanical properties and irregularities of
the terrain. Frequently, locomotion over rough terrain is decomposed into:
(a) perception and planning to reason about terrain conditions, by computing a 
plan that allows the legged system to traverse the terrain toward a goal, and
(b) control that executes the plan while compensating for uncertainties in 
perception, modelling errors, etc. In this work, we focus on generating on-line
and versatile plans for quadruped locomotion over challenging terrain.

In legged motion planning one can compute simultaneously contacts and body
movements, leading to a \textit{coupled motion planning approach}
\cite{Tassa2010}\cite{Mordatch2012a}\cite{Posa2013}\cite{Dai2014}.
This can be posed as a hybrid system or a mode-invariant problem. Such
approaches tend to compute richer motion plans than decoupled motion planners,
especially when employing mode-invariant strategies. Nevertheless, these
approaches are often hard to use in a practical setting. They are usually posed
as non-linear optimization problems such as Mathematical Programming with
Complementary Constraints \cite{Posa2013}, and are computationally expensive. On
the other hand, the legged motion planning problem can be posed as a decoupled
approach that is naturally divided into: motion and contact planning
\cite{Kolter2008}\cite{Vernaza2009}\cite{Kalakrishnan2009}. These approaches
avoid the combinatorial search space at the expense of complexity of locomotion.
A decoupled motion planner has to explore different plans in the space of
feasible movements (state space), which is often defined by physical, stability,
dynamic and task constraints. Nevertheless, the feasibility space is variable
since the stability constraints depend on the kind of movement, e.g. static or
dynamic walking.

\begin{figure}%
	\centering	
	\includegraphics[width=0.9\columnwidth]{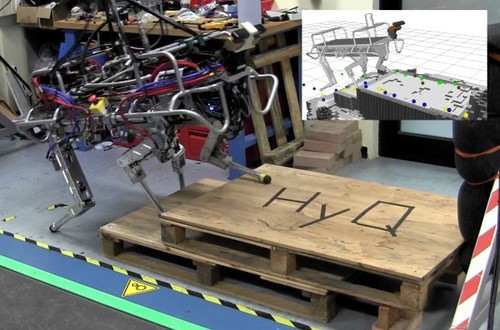}
	\caption{The hydraulically actuated and fully torque-controlled quadruped
	robot HyQ \cite{Semini2011}. The inset plot shows the on-line environment
	perception alongside with the planned footholds and the current on-board robot state
	estimate.}
	\label{fig:hyq}
\end{figure}

The challenge of decoupled planners lies primarily in reducing the computation
time while increasing the complexity of motion generation. To the best of our
knowledge, up to now, decoupled approaches are limited in the versatility of
movements and computation time, for instance \cite{Kolter2008} reduces the
computation time but is still limited to small changes of the robot's yaw
(heading). Therefore, our main contribution is a planning approach that
increases the versatility of plans, based on the definition of footstep search
regions and footstep sequence according to a body action plan.
Our method computes on-line and on-board plans (around \unit[1]{Hz}) using the
incoming perception information on commodity hardware. We evaluate our planning
approach on the Hydraulic Quadruped robot -HyQ- \cite{Semini2011} shown in Fig.
\ref{fig:hyq}.

The rest of the paper is structured as follows: after discussing previous
research in the field of legged motion planning (Section
\ref{sec:related_work}). Section \ref{sec:technical_approach} explains, how
on-line and on-board versatile plans are generated based on a decoupled
planning approach (body action and footstep sequence planners). In Section
\ref{sec:experimental_results} we evaluate the performance of our planning
approach in real-world trials before Section \ref{sec:conclusion}
summarizes this work and presents ideas for future work.

\section{Related Work}\label{sec:related_work}
Motion planning is an important problem for successful legged locomotion in
challenging environments. Legged systems can utilize a variety of dynamic gaits
(e.g. trotting and galloping) in environments with smooth and continuous support
such as flats, fields, roads, etc. Such cyclic gaits often assume that contact
will be available within the reachable workspace at every step. However, for
more complex environments, reactive cyclic gait generation approaches quickly
reach their capabilities as foot placement becomes crucial for the success of
the behavior.

Natural locomotion over rough terrain requires simultaneous computation of 
footstep sequences, body movements and locomotion behaviors (coupled planning)
\cite{Tassa2010}\cite{Mordatch2012a}\cite{Posa2013}\cite{Dai2014}. One of the
main problems with such approaches is that the search space quickly grows and
searching becomes unfeasible, especially for systems that require on-line
solutions. In contrast, we can state the planning and control problem into a set
of sub-problems, following a decoupled planning strategy. For example the body
path and the footstep planners can be separated, thus reducing the search space
for each component \cite{Kolter2008}\cite{Vernaza2009}\cite{Kalakrishnan2009}.
This can reduce the computation time at the expense of limiting the planning
capabilities, sometimes required for extreme rough terrain. There are two main
approaches of decoupled planning: \textit{contact-before-motion}
\cite{Escande2006}\cite{Hauser2009}\cite{Vernaza2009} and
\textit{motion-before-contact} \cite{Zucker2010}\cite{Kolter2008}. These
approaches find a solution in motion space, which defines the possible motion of
the robot.

The motion space of \textit{contact-before-motion} planner lies in a
sub-manifold of the configuration space\footnote{A configuration space defines
all the possible configurations of the robot, e.g. joint positions.} with
certain motion constraints, thus, a footstep constraint switches the motion
space to a lower-dimensional sub-manifold $Q_\sigma$\footnote{Configuration
space of a certain stance $\sigma$.} \cite{Hauser2010}\cite{Escande2013}.
Therefore, these planners have to find a path that connects the initial
$F_\sigma$ and goal $F_{\sigma'}$ feasible regions. Note that the planner must
find transitions between feasible regions and then compute paths between all the
possible stances (feasible regions) that connect $F_\sigma$ and $F_{\sigma'}$,
which is often computationally expensive. On the other hand,
\textit{motion-before-contact} approaches allow us to shrink the search space
using an intrinsic hierarchical decomposition of the problem into; high-level
planning, body-path planning and footstep planning, and low-level planning, CoG
trajectory and foot trajectory generation. These approaches reduce considerably
the search space but can also limit the complexity of possible movements,
nevertheless \cite{Kolter2008}, \cite{Kalakrishnan2010b} and \cite{Winkler2014}
have demonstrated that this can be a successful approach for rough terrain
locomotion. Nonetheless these approaches were tailored for specific types of
trials where goal states are mostly placed in front of the robot. This way most
of these frameworks developed planners that use a fixed yaw and only allow
forward expansion of the planned paths. Our approach takes a step forward in
versatility by allowing motion in all directions and changes in the robot's yaw
(heading) as in real environments this is almost always required.

The planning approach described in this paper decouples the problem into: body
action and footstep sequence planning (\textit{motion-before-contact}). The
decoupling strategy allows us to reduce the computation time, ensures
sub-optimal plans, and computes plans on-line using the incoming information of
the terrain. Our body action planner uses a lattice representation based on body
movement primitives. Compared to previous approaches our planning approach
generates on-line more versatile movements (i.e. forward and backward, diagonal,
lateral and yaw-changing body movements) while also ensure feasibility for
every body action. Both planning and perception are computed on-line and
on-board.

\section{Technical Approach}\label{sec:technical_approach}
Our overall task is to plan on-line an appropriate sequence of footholds
$\mathbf{F}$ that allows a quadruped robot to traverse a challenging terrain
toward a body goal state $(x,y,\theta)$. To accomplish this, we decouple the
planning problem into: body action and footstep sequence planning. The body
action planner searches a bounded sub-optimal solution around a growing
\textit{body-state graph}. Then, the footstep sequence planner selects a
foothold sequence around an action-specific foothold region of each planned
body action.

Clearly, decoupled approaches reduce the combinatorial search space at the
expense of the complexity of locomotion. This limits the versatility of the body
movement (e.g. discretized yaw--changing movements instead of continuous
changes). Nevertheless, our approach manages this trade-off by employing a
lattice representation of potentially feasible body movements and
action-specific foothold regions.

\subsection{Terrain Reward Map}
\begin{figure}%
	\centering
	\includegraphics[width=0.9\columnwidth]{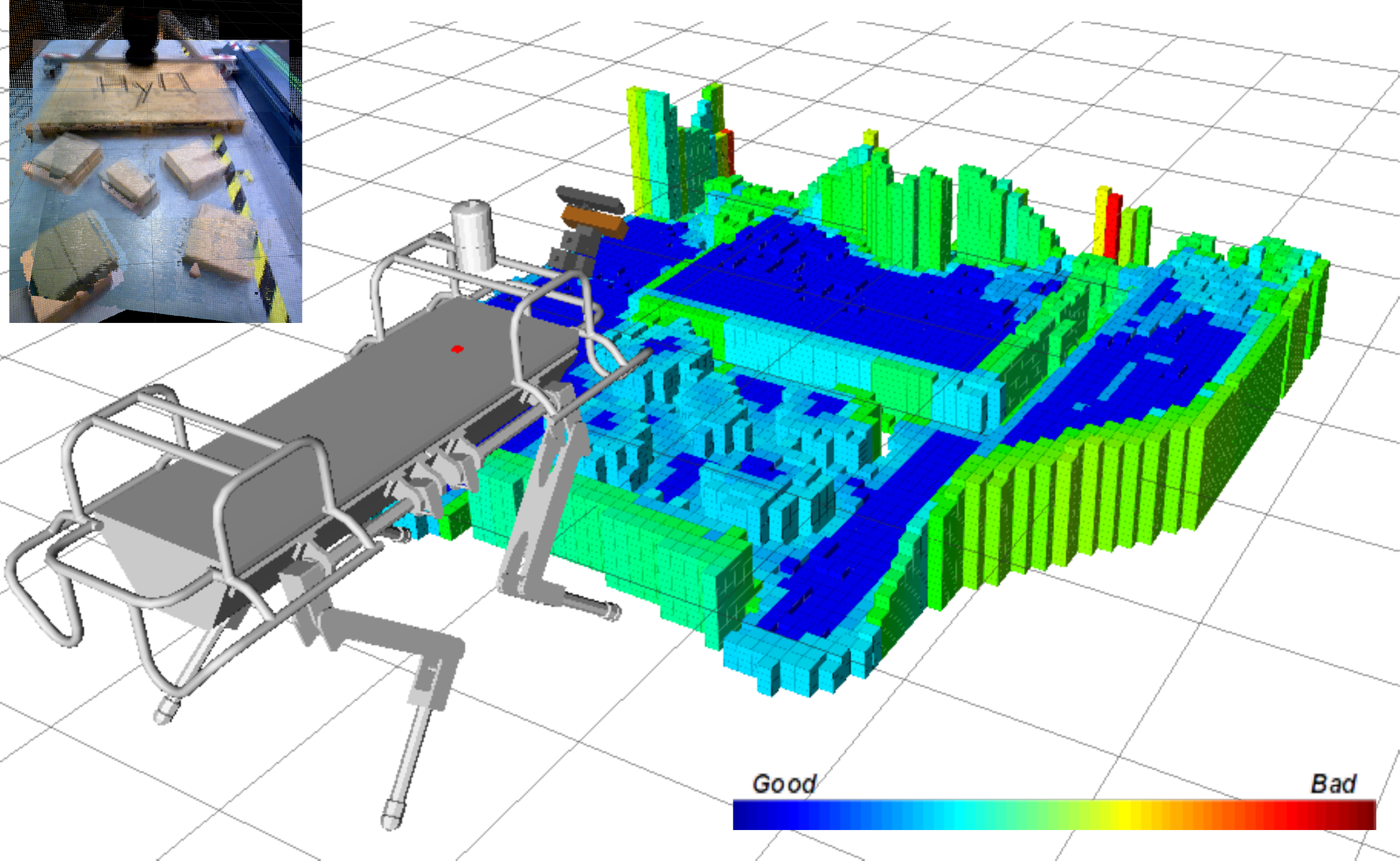}
	\caption{The reward map generation from RGBD (Asus Xtion) camera data. The RBGD
	sensor is mounted on a PTU unit that scans the terrain. An occupancy map is 
	built with a \unit[2]{cm} resolution. Then the
	set of features is computed and the total reward value per pixel of
	\unit[4]{cm} is calculated. In addition, a height map of \unit[2]{cm}
	resolution in $z$ is created. The reward values are represented using a color 
	scale, where blue is the maximum reward and red is the minimum one.}
	\label{fig:reward_map}
\end{figure}

The terrain reward map quantifies how desirable it is to place a foot at a
specific location. The reward value for each cell in the map is computed using
geometric terrain features as in \cite{Winkler2014}. Namely we use the
\textit{standard deviation of height values}, the \textit{slope} and the
\textit{curvature} as computed through regression in a 6cm$\times$6cm window
around the cell in question; the feature are computed from a voxel model (2
cm resolution) of the terrain. We incrementally compute the reward map based on
the aforementioned features and recompute locally whenever a change in the map
is detected. For this we define an area of interest around the robot of
2.5m$\times$5.5m that uses a cell grid resolution of \unit[4]{cm}. The cost
value for each cell of the map is computed as a weighted linear combination of
the individual feature $R(x,y) = \mathbf{w}^T \mathbf{r}(x,y)$, where
$\mathbf{w}$ and $\mathbf{r}(x,y)$ are the the weights and features values,
respectively. Figure \ref{fig:reward_map} shows the generation of the reward map
from the mounted RGBD sensor. The reward values are represented using a color
scale, where blue is the maximum reward and red is the minimum one.

\subsection{Body Action Planning}
A body action plan over challenging terrain take into consideration the
wide range of difficulties of the terrain, obstacles, type of action, potential
shin collisions, potential body orientations and kinematic reachability. Thus,
given a reward map of the terrain, the body action planner computes a sequence
of body actions that maximize the \textit{cross-ability} of the sub-space of
candidate footsteps. The body action plans are computed by searching over a
\textit{body-state graph} that is built using a set of predefined body movements
primitives. The body action planning is described in Algorithm
\ref{alg:body_action_planning}.

\begin{algorithm}
\begin{algorithmic}\\
 	\caption{Computes a set of body action over a challenging terrain using an
 	ARA$^*$ search.}
	\label{alg:body_action_planning}
	\State{\textbf{Data:} Inflation parameter $\epsilon$, time of computation
	$t_c$}
	\State{\textbf{Result:} a body action plan
	$\mathbf{Q}=[\mathbf{q}^0,\mathbf{q}^1,\cdots,\mathbf{q}^l]$}
 	\Function{computeBodyActionPlan}{$\mathbf{s}_{start},\mathbf{s}_{goal}$}
	\State{set $\epsilon$ and computation time $t_c$}
	\State{$g(\mathbf{s}_{start})=0; g(\mathbf{s}_{goal})=\infty$}
	\State{OPEN $=\{\mathbf{s}_{start}\}$; INCONS $=$ CLOSED$=\emptyset$}
	\While{($\epsilon\geq 1$ and $t_c<t$)}
		\State{decrease $\epsilon$}
		\State{OPEN $=$ OPEN $\cup$ INCONS; CLOSED $=\emptyset$}
		\State{// fval$(\mathbf{s})=g(\mathbf{s})+\epsilon h(\mathbf{s})$}
		\While{(fval$(\mathbf{s}_{goal})>$ minimum fval in OPEN)}
			\State{remove $\mathbf{s}$ with minimum fval from OPEN}
			\State{insert $\mathbf{s}$ into CLOSED}
			\State{generate action($\mathbf{s}$) from body primitives}
			\ForAll{$\mathbf{u}\in$ action($\mathbf{s}$)}
				\State{compute $\mathbf{s'}$ given $\mathbf{u}$}
				\State{compute the footstep regions given $\mathbf{s'}$}
				\State{compute $c_{body}(\mathbf{s},\mathbf{s'})$ from footstep
			regions}
				\If{$g(\mathbf{s'})>g(\mathbf{s})+c_{body}(\mathbf{s},\mathbf{s'})$}
					\State{$g(\mathbf{s'})=g(\mathbf{s})+c_{body}(\mathbf{s},\mathbf{s'})$}
					\State{add $\mathbf{s}\mapsto\mathbf{s'}$ to policy
					$\mathbf{s'}=\pi(\mathbf{s})$}
					\If{$\mathbf{s'}$ is not in CLOSED}
						\State{insert ($\mathbf{s'}, $fval($\mathbf{s'}$)) into OPEN}
					\Else
						\State{insert ($\mathbf{s'}$, fval($\mathbf{s'}$)) into INCONS}
					\EndIf
				\EndIf
			\EndFor
		\EndWhile
	\EndWhile
	\State{reconstruct body action plan $\mathbf{Q}$ from
	$\pi(\cdot)$}
	\State{\Return{$\mathbf{Q}$}}
  	\EndFunction
\end{algorithmic}
\end{algorithm}

\subsubsection{Graph construction}
We construct the \textit{body-state graph} using a lattice-based adjacency
model. Our lattice representation is a discretization of the configuration space
into a set of feasible body movements, in which the feasibility depends on the
selected sequence of footholds around the movement. The \textit{body-state
graph} represents the transition between different body-states (nodes) and it is
defined as a tuple, $\mathcal{G} = (\mathcal{S},\mathcal{E})$, where each node
$\mathbf{s} \in \mathcal{S}$ represents a \textit{body-state} and each edge
$\mathbf{e} \in \mathcal{E} \subseteq \mathcal{S} \times \mathcal{S}$ defines a
potential feasible transition from $\mathbf{s}$ to $\mathbf{s'}$. A sequence of
body-states (or body poses $(x,y,\theta)$) approximates the body trajectory that
the controller will execute. An edge defines a feasible transition (body action)
according to a set of body movement primitives. The body movement primitives
are defined as body displacements (or body actions), which ensure feasibility
together with a feasible footstep region. A feasible footstep region is defined
according to the body action.

\begin{figure}%
	\centering
	\includegraphics[width=0.9\columnwidth]{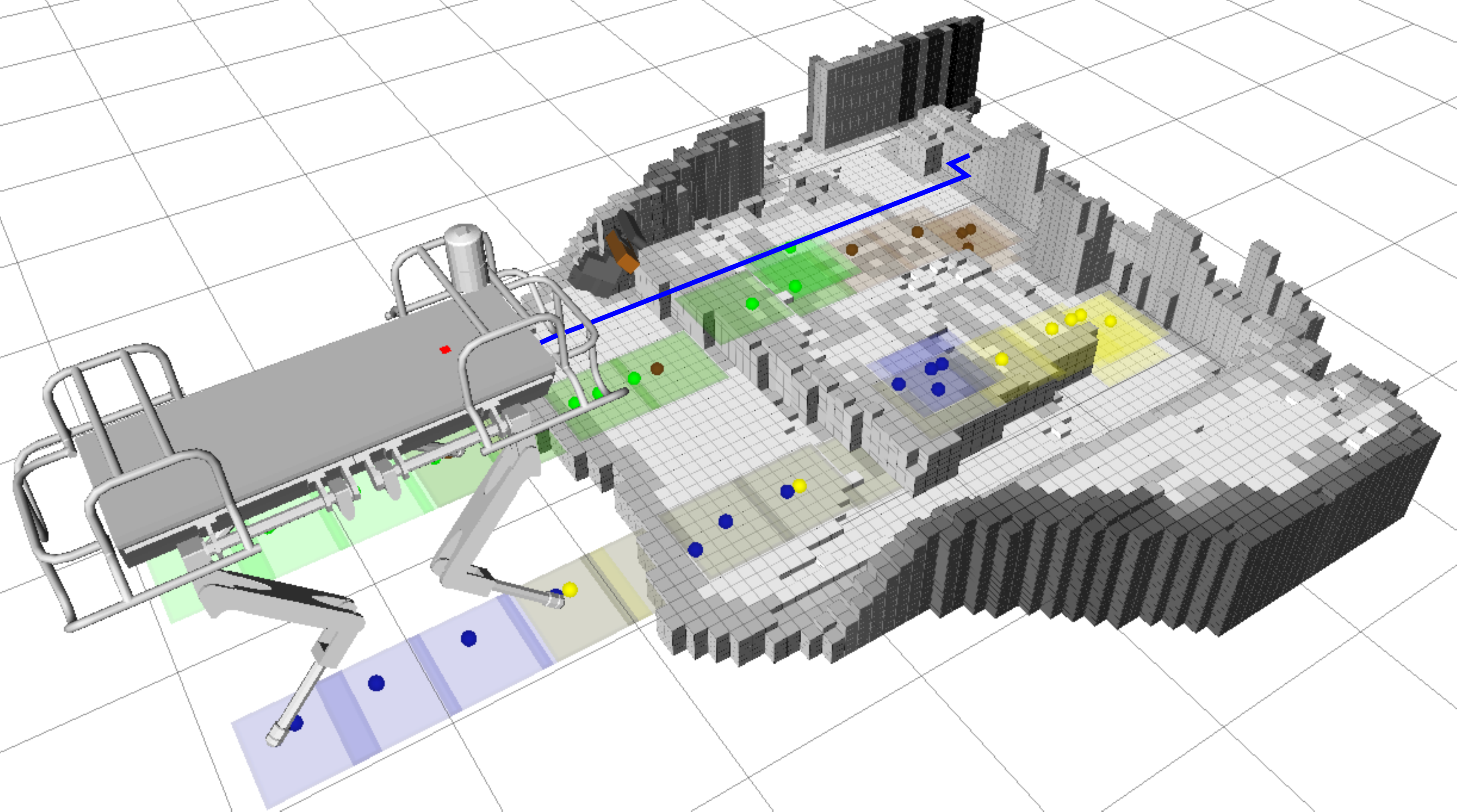}
	\caption{Illustration of graph construction of the least-cost path. The
	associated cost of every transition $c_{body}(\mathbf{s},\mathbf{s'})$ is
	computed using the footstep regions. For every expansion, the footstep
	regions of LF (brown squares), RF (yellow squares), LH (green squares) and RH
	(blue squares) are computed according to a certain body action.}
	\label{fig:graph_construction}
\end{figure}

Given a body-state query, a set of successor states are computed using a set of
predefined body movement primitives. A predefined body movement primitive
connects the current body state $\mathbf{s}=(x,y,\theta)$ with the successor
body state $\mathbf{s'}=(x',y',\theta')$. The graph $\mathcal{G}$ is dynamically
constructed due to the associated cost of transition
$c_{body}(\mathbf{s},\mathbf{s'})$ depends on the current and next states (or
current state and action). In fact, the feasible regions of foothold change
according to each body action, which affect the value of the transition cost.
Moreover, an entire graph construction could require a greater memory
pre-allocation and computation time than available (on-board computation).
Figure \ref{fig:graph_construction} shows the graph construction for the body
action planner. The associated cost of every transition
$c_{body}(\mathbf{s},\mathbf{s'})$ is computed using the footstep regions. These
footstep regions depend on the body action in such a way that they ensure
feasibility of the plan, as is explained in \ref{sec:ensuring_feasibility} subsection.
For every expansion, the footstep regions of Left-Front LF (brown squares),
Right-Front RF (yellow squares), Left-Hind LH (green squares) and Right-Hind RH
(blue squares) legs are computed given a body action.

The resulting states $\mathbf{s'}$ are checked for body collision with
obstacles, using a predefined area of the robot, and invalid states are
discarded. For legged locomotion over challenging terrain, obstacles are defined
as unfeasible regions to cross, e.g. a wall or tree. In our case, the obstacles
are detected when the height deviation w.r.t. the estimated plane of the ground
is larger than the kinematic feasibility of the system in question (HyQ). We
build the obstacle map with \unit[8]{cm} resolution, otherwise it would be
computationally expensive due to the fact that it would increase the number of
collision checks. In fact, the body collision checker evaluates if there are
body cells in the resulting state that collide with an obstacle.

\subsubsection{Body cost}
The body cost describes how desirable it is to cross the terrain with a certain
body path (or body actions). This cost function is designed to maximize the
cross-ability of the terrain while minimizing the path length. In fact, the body
cost function $c_{body}$ is a linear combination of: terrain cost $c_t$, action
cost $c_a$, potential shin collision cost $c_{pc}$, and potential body
orientation cost $c_{po}$. The cost of traversing any transition between the
body state $\mathbf{s}$ and $\mathbf{s'}$ is defined as follows:
\begin{eqnarray}\label{eq:body_cost}\nonumber
	c_{body}(\mathbf{s},\mathbf{s'}) = w^b_t c_t(\mathbf{s}) + w_a
	c_a(\mathbf{s},\mathbf{s'}) \\
	+ w_{pc} c_{pc}(\mathbf{s}) + w_{po} c_{po}(\mathbf{s})
\end{eqnarray}
where $w_i$ are the respective weight of the different costs.

\begin{figure}[b]
	\centering
	\includegraphics[width=0.7\columnwidth]{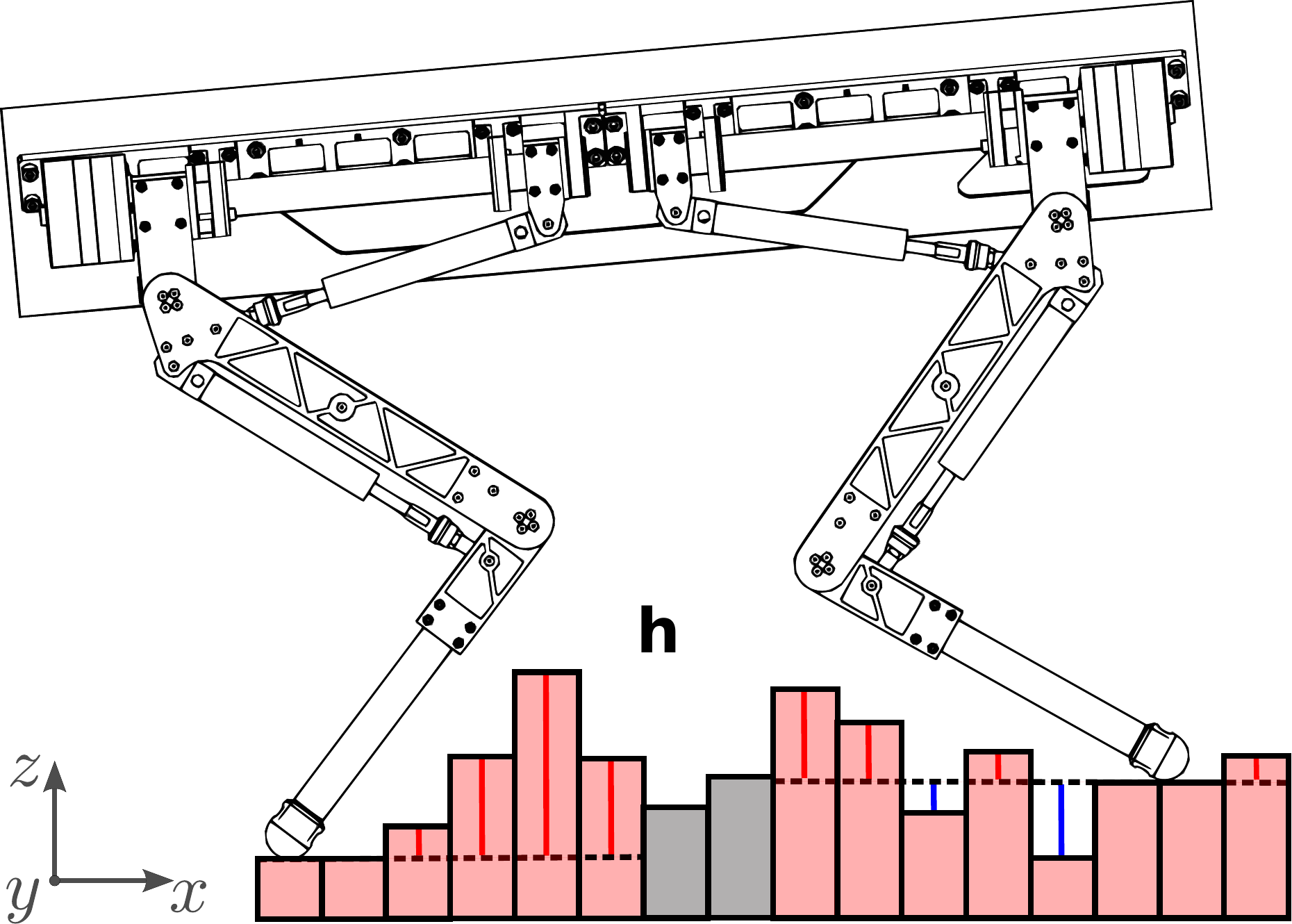}
	\caption{Computation of the potential shin collision. The potential shin
	collisions are detected around a predefined shin collision region (red bars).
	The computed cost is proportional to the height defined (red lines) around the
	footstep plane (dashed black lines), where blue height differences do not
	contribute to the cost.}
	\label{fig:shin_collision}
\end{figure}

For a given current body state $(x,y,\theta)$, the terrain cost $c_t$ is
evaluated by averaging the best $n$-footsteps terrain reward values around the
foothold regions of a nominal stance (nominal foothold positions). The action
cost $c_{a}$ is defined by the user according to the desirable actions of the
robot, e.g. it is preferable to make diagonal body movements than lateral ones.
The potential leg collision cost $c_{pc}$ is computed by searching potential
obstacles in the predefined workspace region of the foothold, e.g. near to the
shin of the robot. In fact, a potential shin collision is detected around a
predefined shin collision region, which depends on the configuration of each leg
as shown in Figure \ref{fig:shin_collision}. This cost is proportional to the
height defined around the footstep plane, where red bars represent collision
elements. Finally, the potential body orientation $c_{po}$ is estimated by
fitting a plane in the possible footsteps around the nominal stance for each
leg.

\subsubsection{Reducing the search space}
Decoupling the planning problem into body action and footstep planning avoids
the combinatorial search space. This allows us to compute plans on-line, and
moreover, to develop a closed-loop planning approach that can deal with changes
in the environment. Nevertheless the reduced motion space might span
infeasible regions due to the strong coupling between the body action and
footstep plans. A strong coupling makes that a feasible movement depends on the
whole plan (body action and footstep sequence). In contrast to previous
approaches
\cite{Kolter2008}\cite{Vernaza2009}\cite{Kalakrishnan2010b}\cite{Zucker2010}\cite{Winkler2014},
our planner uses a lattice-based representation of the configuration space. The
lattice representation uses a set of predefined body movement primitives that
allows us to apply a set of rules that ensure feasibility in the generation of
on-line plans.

The body movement primitives are defined as 3D \textit{motion-actions} $(x,y,\theta)$ of
the body that discretize the search space. We implement a set of different
movements (\textit{body movement primitives}) such as: forward and backward, diagonal,
lateral and yaw-changing movements.

\subsubsection{Ensuring feasibility}\label{sec:ensuring_feasibility}
\begin{figure}[b]
	\centering
 	\includegraphics[width=0.85\columnwidth]{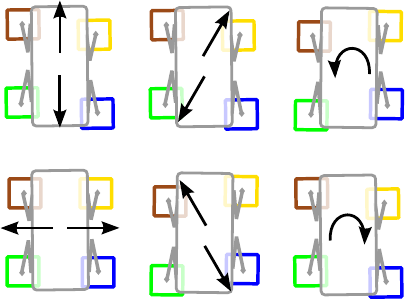}
	\caption{Different footstep search regions according to the body action. These
	footstep regions ensure the feasibility of the plan. Moreover, it increase the
	triangular support areas, improving the execution of the whole-body plan. The
	brown, yellow, green and blue squares represent the footstep search regions for
	LF, RF, LH and RH feet, respectively.}
	\label{fig:footstep_regions}
\end{figure}

Our lattice representation allows the robot to search a body action plan
according to a set of predefined body movement primitives. A predefined body
movement primitive cannot guarantee a feasible plan due to the mutual dependency
of movements and footsteps. For instance, the feasibility of clockwise movements
depends on the selected sequence of footsteps, and can only be guaranteed in a
specific footstep region. In fact, we exploit this characteristic in order to
ensure feasibility in each body movement primitive. These footstep regions
increase the support triangle areas given a certain body action, improving the
execution of the whole-body plan. This strategy also guarantees that the body
trajectory generator can connect the support triangles (support triangle
switching) \cite{Kalakrishnan2010b}. Figure \ref{fig:footstep_regions}
illustrates the footstep regions according to the type of body movement
primitives. These footstep regions are tuned to increase the triangular support
areas, improving the execution of the whole-body plan.

\subsubsection{Heuristic function}
The heuristic function guides the search to promising solutions, improving the
efficiency of the search. Heuristic-based search algorithms require that the
heuristic function is admissibile and consistent. Most of the heuristic
functions estimate the cost-to-go by computing the Euclidean distance to the goal.
However, these kind of heuristic functions do not consider the terrain
conditions. Therefore, we implemented a terrain-aware heuristic function that 
considers the terrain conditions.

The terrain-aware heuristic function computes the estimated cost-to-go by
averaging the terrain cost locally and estimating the number of footsteps to the
goal:
\begin{equation}\label{eq:heuristic_function}
	h(\mathbf{s}) = -\bar{R}\mathcal{F}(\|\mathbf{g}-\mathbf{s}\|)
\end{equation}
where $\bar{R}$ is the average reward (or cost) and
$\mathcal{F}(\|\mathbf{g}-\mathbf{s}\|)$ is the function that estimates the
number of footsteps to the goal.

\subsubsection{Ensuring on-line planning}
Open-loop planning approaches fail to deal adequately with changes in the
environment. In real scenarios, the robot has a limited range of perception that
makes open-loop planning approaches unreliable. Closed-loop planning considers
the changes in the terrain conditions, and uses predictive terrain conditions
for non-perceived regions, improving the robustness of the plan. Dealing with
re-planning and updating of the environment information requires that the
information is managed in an efficient search exploration of the state space.
We manage to reduce the computation time of building a reward map by computing
the reward values from a voxelized map of the environment. Additionally, we
(re-)plan and update the information using ARA$^*$ \cite{Likhachev2004}. ARA$^*$
ensures provable bounds on sub-optimality, depending on the definition of the
heuristic function. On the other hand, the terrain-aware heuristic function
guides the search according to terrain conditions.

\subsection{Footstep Sequence Planning}
Given the desirable body action plan,
$\mathbf{Q}=(\mathbf{x}_d,\mathbf{y}_d,\boldsymbol{\theta}_d)$, the footstep
sequence planner computes the sequence of footholds that reflects the intention of the
body action. A local greedy search procedure selects the optimal footstep
target. A (locally optimal) footstep is selected from defined footstep search
regions by the body action planner. For every body action, the footstep planner
finds a sequence of the 4-next footsteps. The footstep sequence planning is
described in Algorithm \ref{alg:footstep_sequence_planning}.

\begin{algorithm}
\begin{algorithmic}\\
 	\caption{Computes a footstep sequence given a body action plan.}
	\label{alg:footstep_sequence_planning}
	\State{\textbf{Data:} footstep horizon F}
	\State{\textbf{Result:} a footstep sequence \textbf{F}}
	\Function{computeFootstepSequence}{$\mathbf{Q}$}
	\State{set horizon F}
	\For{$\mathbf{q}_{i=0:F}\in\mathbf{Q}$}
		\For{i=0:3}
			\State{compute the sequence of footstep $e$ given $\mathbf{q}_i$}
			\State{generate possible FOOTSTEP($e,\mathbf{q}_i$)}
			\State{compute the greedy solution $\mathbf{f}_{min}$}
			\State{add $\mathbf{f}_{min}$ to $\mathbf{F}$}
		\EndFor
	\EndFor
	\State{\Return{$\mathbf{F}$}}
	\EndFunction
\end{algorithmic}
\end{algorithm}

\subsubsection{Footstep cost}
The footstep cost describes how desirable is a foothold target,
given a body state $\mathbf{s}$. The purpose of this cost function is to
maximize the locomotion stability given a candidate set of footsteps. The footstep cost
$c_{footstep}$ is a linear combination of: terrain cost $c_t$, support triangle
cost $c_{st}$, shin collision cost $c_c$ and body orientation cost $c_o$.
The cost of certain footstep $\mathbf{f}^e$ is defined as follows:
\begin{eqnarray}\label{eq:footstep_cost}\nonumber
	c_{footstep}(\mathbf{f}^e) = w^f_t c_t(\mathbf{f}^e) + w_{st}
	c_{st}(\mathbf{f}^e) \\
	+ w_{c} c_{c}(\mathbf{f}^e) + w_{o} c_{o}(\mathbf{f}^e)
\end{eqnarray}
where $\mathbf{f}^e$ defines the Cartesian position of the foothold target $e$
(foot index). We consider as end-effectors: the LF, RF, LH and RH feet of HyQ.

The terrain cost $c_t$ is computed from the terrain reward value of the
candidate foothold, i.e. using the terrain reward map (see Figure
\ref{fig:reward_map}). The support triangle cost $c_{st}$ depends on the inradii
of the triangle formed by the current footholds and the candidate one. As in the
body cost computation, we use the same predefined collision region around the
candidate foothold. Finally, the body orientation cost $c_{o}$ is computed using
the plane formed by the current footsteps and the candidate one. We calculate
the orientation of the robot from this plane.

\section{Experimental Results}\label{sec:experimental_results}
The following section describes the experiments conducted to validate the
effectiveness and quantify the performance of our planning approach.

\subsection{Experimental Setup}
We implemented and tested our approach with HyQ, a Hydraulically actuated
Quadruped robot developed by the Dynamic Legged Systems (DLS) Lab
\cite{Semini2011}. HyQ roughly has the dimensions of a medium-sized dog, weight
\unit[90]{kg}, is fully-torque controlled and is equipped with precision joint
encoders, a depth camera (Asus Xtion) and an Inertial Measurement Unit (IMU). We
used an external state measuring system, Vicon marker-based tracker system. The
perception and planning is done on-board in an i7/2.8GHz PC, and all
computations of the real-time controller are done using a PC104 stack.

\begin{figure}%
	\centering
 	\includegraphics[width=0.95\columnwidth]{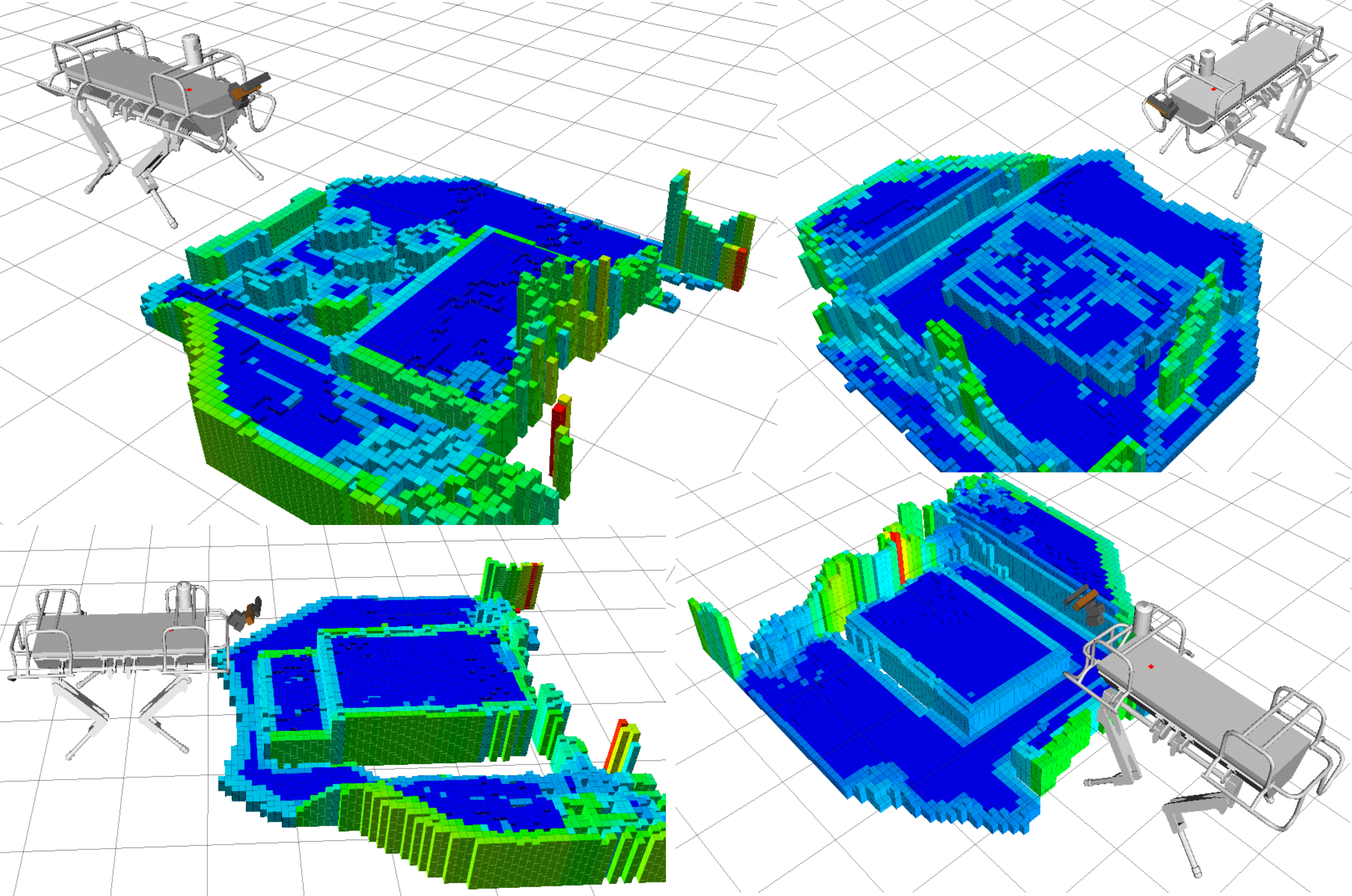}
	\caption{The planning benchmarks used to analyse 
	the quality of the produced plans. Top left: stepping stones; top right:
	pallet; bottom left: stair; bottom right: gap.}
	\label{fig:planning_benchmarks}
\end{figure}

The first set of experiments are designed to test the capabilities of our
motion planner. We use a set of benchmarks of realistic locomotion scenarios
(see Figure \ref{fig:planning_benchmarks}): stepping stones, pallet, stair and
gap. In these experiments, we compared the cost, number of expansions and
computation time of ARA$^*$ against A$^*$ \cite{Likhachev2004} using our lattice
representation. The results are based on 9 predefined goal locations. Second
experiment, the robot must plan on-line with dynamic changes in the terrain.
Third experiment, the next experiment we show the on-line planning and
perception results. Finally, we validate the performance of our planning
approach by executing in HyQ.

\subsection{Results and Discussions}
\subsubsection{Initial plan results}
\begin{figure}[!b]
	\centering
 	\includegraphics[width=0.78\columnwidth]{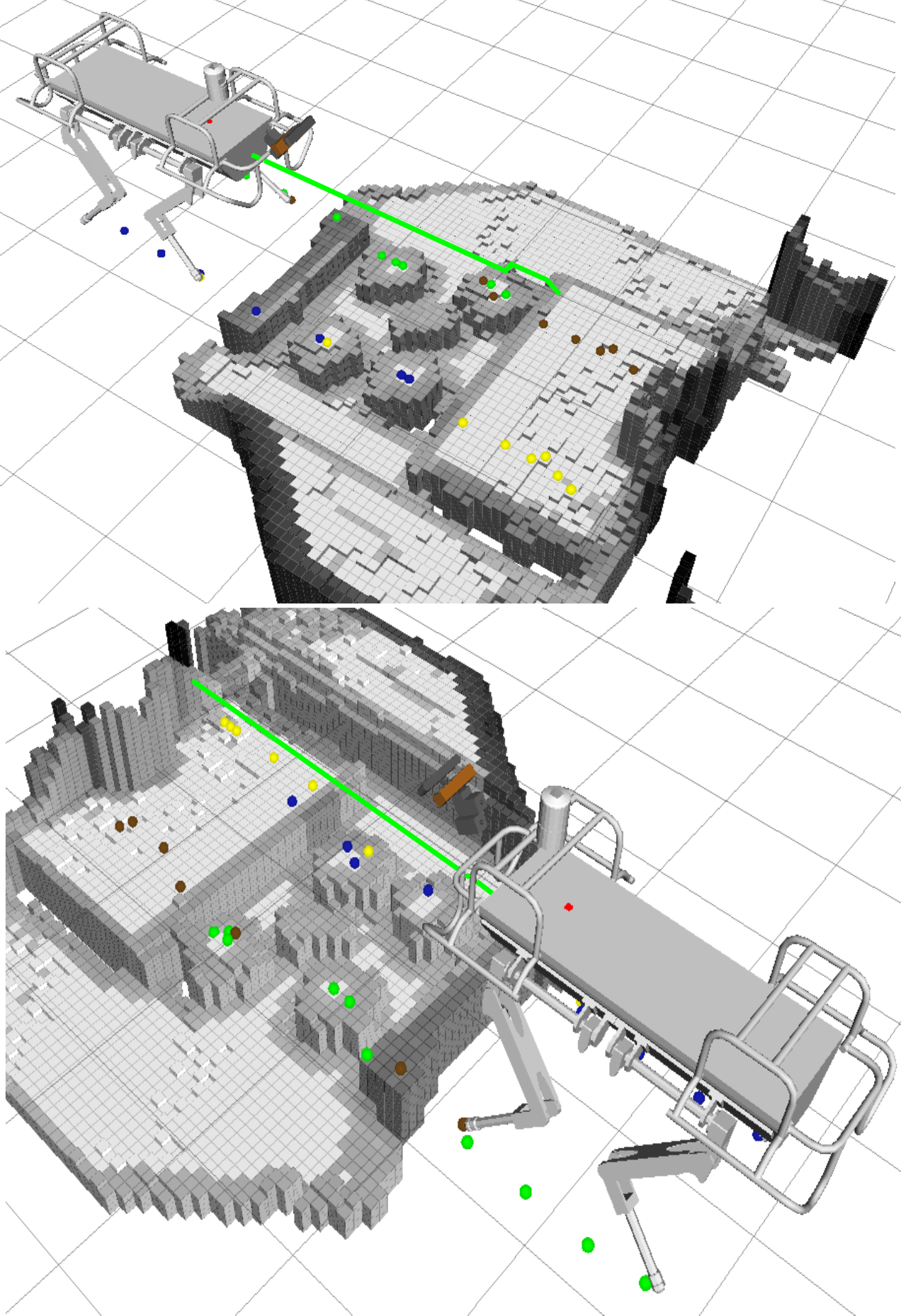}
	\caption{Comparison of the body action (green line) and footstep sequence
	computed plan A$^*$ and	ARA* given reward map (grey scale). The brown, yellow,
	green and blue points represent the planned footsteps of the LF, RF, LH and RH
	feet, respectively. The top image shows the computed plan by A$^*$, and the
	bottom one the results of ARA$^*$.}
	\label{fig:compared_planning}
\end{figure}

The stepping stones, pallet, stair and gap experiments (see Figure
\ref{fig:planning_benchmarks}) show the initial plan quality (see Table
\ref{tab:results}) of our approach using A$^*$ and ARA$^*$. To this end, we plan
a set of body actions and footstep sequences for 9 predefined goal locations, 
approximately \unit[2.0]{m} away from the starting position, and compare the 
cost and number of expansions of the body action path, and the planning time of
ARA$^*$ against A$^*$. Three main factors contribute to the decreased planning
time while maintaining the quality of the plan:

First, the lattice-based representation (using body movement primitives)
considers versatile movements in the sense that it allows us to reduce the
search space around feasible regions (feasible motions) according a certain
body action, in contrast to grid-based approaches that ensure the feasibility by
applying rules that are agnostic to body actions. Second, our terrain-aware
heuristic function guides the tree expansion according to terrain conditions in
contrast to a simple Euclidean heuristic. Finally, the ARA* algorithm 
implements a search procedure that guarantees bounded sub-optimality in the
solution given a proper heuristic function \cite{Likhachev2004}. Figure
\ref{fig:compared_planning} shows the initial plan of A* and ARA*.

\begin{table}[!b]
	\centering
	\caption{Cost of the plan (Cost), number of expansions (Exp.) and
	computation time (Time) averaged over 9 trials of A$^*$ and ARA$^*$.}
	\begin{tabular}{@{} lrrrrrr @{}}    
	\toprule 
	&\multicolumn{3}{c}{A*} & \multicolumn{3}{c}{ARA*} \\
	\cmidrule{2-4}\cmidrule{5-7}
	\emph{Terrain}	& Cost  & Exp. & Time [s]  & Cost  & Exp. & Time [s] \\
	\midrule
 	S. Stones	  	& 364.7 & 3191 & 428.7 & 597.4 & 12.1 & 1.59 \\ 
 	Pallet 			& 269.2 &  270 & 271.2 & 587.7 & 12.7 & 1.74 \\ 
 	Stair  			& 306.1 &  306 & 308.1 & 646.4 & 12.7 & 1.55 \\
 	Gap				& 325.6 &  326 & 327.6 & 647.0 & 13.0 & 2.01 \\
	\bottomrule
	\end{tabular}
	\label{tab:results}
\end {table}

\subsubsection{On-line planning and perception}
Using a movement primitive-based lattice search reduces the size of the search
space significantly, leading to responsive planning and replanning. In our
experimental trials, we choose a lattice graph resolution (discretization) of
\unit[4]{cm} for $x/y$ and 1.8$^\circ$ for $\theta$ and goal state is never more
that \unit[5]{m} away from the robot. In these trials, the plan/replan frequency
is approximately \unit[0.5]{Hz}.

\begin{figure}%
	\centering
 	\includegraphics[width=0.91\columnwidth]{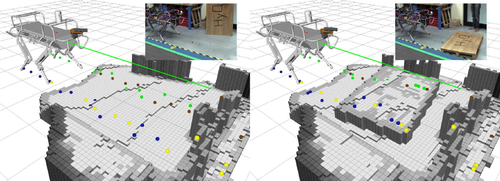}
	\caption{(Re-)planning and perception on-board. The left image presents the
	plan for a flat terrain, then, the right image reflects the re-planning and
	updating of the reward map (grey scale map). The green, brown, blue and yellow
	points represent the planned footholds of the LH, LF, RH, RF legs,
	respectively. The green line represents the body path according to action
	plan.}
	\label{fig:replanning}
\end{figure}

The efficient occupancy grid-based mapping allows us to incrementally build up
the model of the environment and focus our computations to the area of interest
around the robot body, generating plans quickly. This allows us to locally
update the computed reward map and incrementally build the reward map as the
robot moves with an average response frequency of \unit[2]{Hz} as is shown in
Figure \ref{fig:replanning}.

\begin{figure*}[t]
	\centering
	\includegraphics[width=0.91\textwidth]{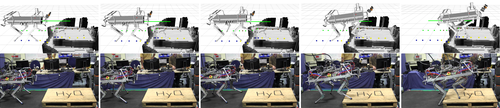}
	\caption{Snapshots of pallet trial used to evaluate the performance
	of our planning approach. From top to bottom: planning and terrain reward map;
	execution of the plan with HyQ.}
	\label{fig:vidsnaps}
\end{figure*}

\subsubsection{Planning and execution}
We generate swift and natural dynamic whole-body motions from an n-step
lookahead optimization of the body trajectory that uses a dynamic stability
metric, the Zero Moment Point (ZMP). A combination of floating-base inverse
dynamics and virtual model control accurately executes such dynamic whole-body
motions with an actively compliant system \cite{Winkler2015}. Figure
\ref{fig:vidsnaps} shows the execution of an initial plan. Our locomotion system
is robust due to a combination of on-line planning and compliance control.

\section{Conclusion}\label{sec:conclusion}
We presented a planning approach that allows us to plan on-line, versatile 
movements. Our approach plans a sequence of footsteps given a body
action plan. The body action planner guarantees a bounded sub-optimal plan given
a set of predefined body movement primitives (lattice representation). In
general, to generate versatile movement one has to explore a considerable region
of the motion space, making on-line planning a computationally challenging task, 
especially in practical applications.
Here, we reduce the
exploration by ensuring the feasibility of every possible body action. In fact,
we define the footstep search regions and footstep sequence based on the body
action. We showed how our planning approach computes on-line plans given
incoming terrain information. Various real-world experimental trials demonstrate
the capability of our planning approach to traverse challenging terrain.

\bibliographystyle{IEEEtran}
\bibliography{ref/planning,ref/control,ref/learning,ref/locomotion,ref/perception,ref/robot}

\end{document}